\documentclass[letterpaper]{article} 
\usepackage{aaai23}  
\usepackage{times}  
\usepackage{helvet}  
\usepackage{courier}  
\usepackage[hyphens]{url}  
\usepackage{graphicx} 
\usepackage{natbib}  
\usepackage{caption} 
\usepackage{amsmath,amssymb,amsfonts}
\usepackage{array}
\usepackage{textcomp}

\usepackage{stfloats}
\usepackage{url}
\usepackage{breakurl}
\usepackage{textcomp}
\usepackage{booktabs}

\usepackage{longtable}
\usepackage{multirow}
\usepackage{multicol}
\usepackage{footnote}
\usepackage{mathrsfs}
\usepackage{threeparttable}

\usepackage{algorithm}
\usepackage{algorithmic}

\usepackage{newfloat}
\usepackage{listings}
\DeclareCaptionStyle{ruled}{labelfont=normalfont,labelsep=colon,strut=off} 
\floatstyle{ruled}
\newfloat{listing}{tb}{lst}{}
\floatname{listing}{Listing}
\title{Causal Conditional Hidden Markov Model for Multimodal Traffic Prediction}
\author {
    Yu Zhao\textsuperscript{\rm 1},
    Pan Deng\textsuperscript{\rm 1}\thanks{Corresponding author.},
    Junting Liu\textsuperscript{\rm 1},
    Xiaofeng Jia\textsuperscript{\rm 2},
    Mulan Wang\textsuperscript{\rm 1}
}
\affiliations {
    \textsuperscript{\rm 1} Beihang University, Beijing, 100191, China.\\
    \textsuperscript{\rm 2} Beijing Big Data Centre, Beijing, 100024, China.\\
     \{iyzhao, pandeng, liujunting, wangmulan\}@buaa.edu.cn,   jiaxf@jxj.beijing.gov.cn
}

\usepackage{bibentry}

\begin{document}

\maketitle

\begin{abstract}
Multimodal traffic flow can reflect the health of the transportation system, and its prediction is crucial to urban traffic management. Recent works overemphasize spatio-temporal correlations of traffic flow, ignoring the physical concepts that lead to the generation of observations and their causal relationship. Spatio-temporal correlations are considered unstable under the influence of different conditions, and spurious correlations may exist in observations. In this paper, we analyze the physical concepts affecting the generation of multimode traffic flow from the perspective of the observation generation principle and propose a Causal Conditional Hidden Markov Model (CCHMM) to predict multimodal traffic flow. In the latent variables inference stage, a posterior network disentangles the causal representations of the concepts of interest from conditional information and observations, and a causal propagation module mines their causal relationship. In the data generation stage, a prior network samples the causal latent variables from the prior distribution and feeds them into the generator to generate multimodal traffic flow. We use a mutually supervised training method for the prior and posterior to enhance the identifiability of the model. Experiments on real-world datasets show that CCHMM can effectively disentangle causal representations of concepts of interest and identify causality, and accurately predict multimodal traffic flow.
\end{abstract}

\section{Introduction}
\label{Introduction}
Urban transportation systems are generally multimodal in nature, consisting of several interconnected subsystems representing different modes of transportation, such as bike, taxi, bus and car. They aim to meet diverse travel demands and provide residents with a variety of travel options\cite{liang2021joint}. Multimodal traffic flow can reflect the health of the transportation system. Urban traffic managers can formulate corresponding management strategies according to the traffic flow in different environments to improve the smoothness of urban operation. Therefore, multimodal traffic flow prediction is a key part of urban traffic management, providing important data support for traffic guidance \cite{liang2021joint}.

\begin{figure}
    \centering
    \includegraphics[width=0.4\textwidth]{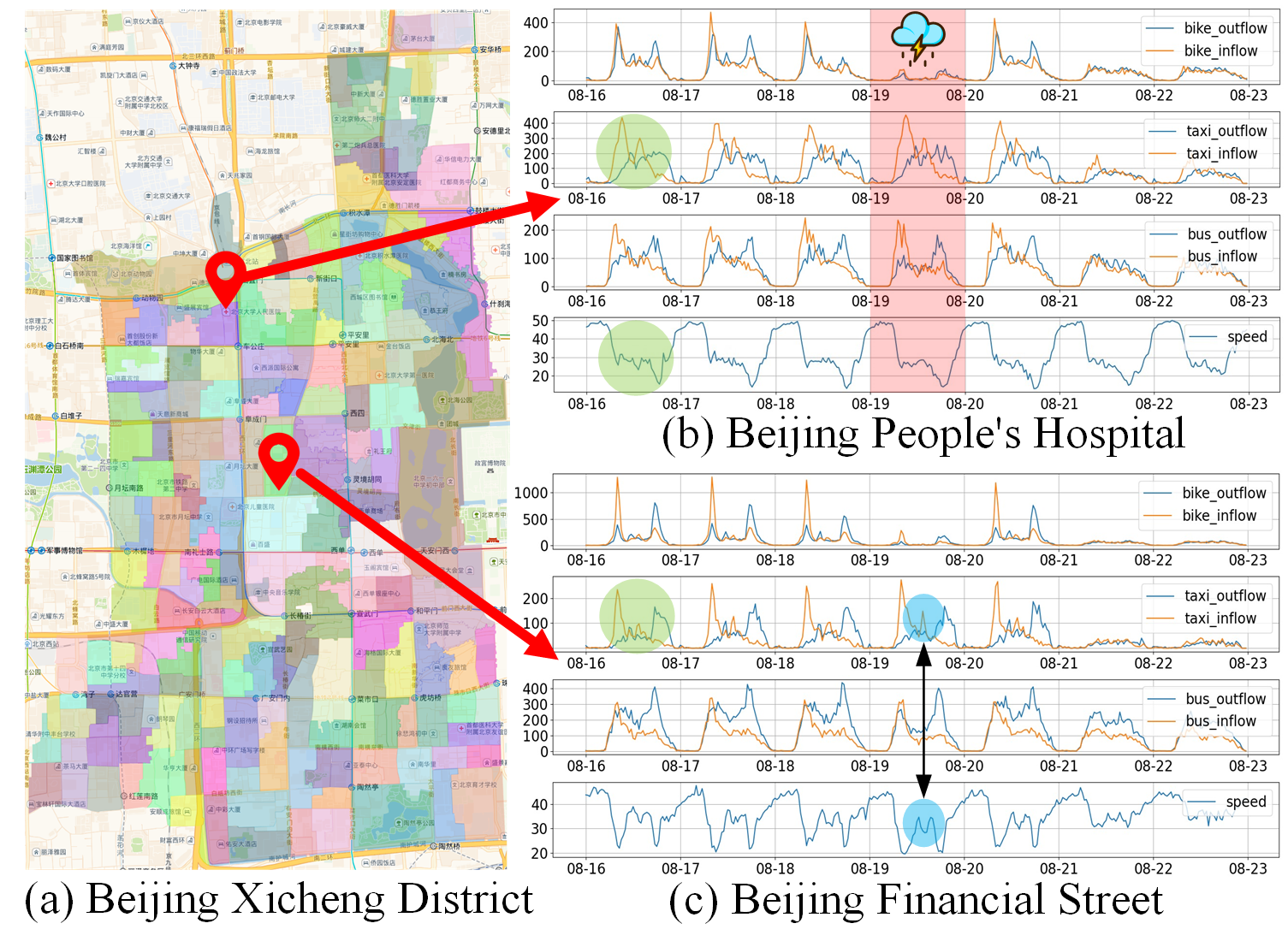}
    \caption{Multimodal traffic flow in different regions.}
    \label{fig:1}
\end{figure}

Most methods only predict a certain traffic flow (e.g., taxi demand or speed)\cite{bai2020adaptive, li2021dynamic, guo2021hierarchical, wu2020connecting, ye2021coupled, han2021dynamic}. They are only partial observations of the traffic system and cannot truly reflect the real situation in real-world scenarios. In contrast, the existing multimodal traffic prediction methods often take different traffic flows as the channel expansion of input data\cite{IOT2021,partc2021,zhou2021modeling, liang2021fine}, or integrate the feature representation of different flows in the model\cite{ye2019co, deng2021pulse}. They implicitly extract the so-called spatio-temporal correlations while lacking the description of causality. However, more input information cannot improve the prediction ability of the model. Instead, it will introduce a large number of confounding factors and extract spurious correlations in observations\cite{comprehensive2021towards, liu2021learning, deng2021comprehensive}, resulting in the decline of model performance.

Nowadays, traffic flow prediction method overemphasizes spatio-temporal correlations of traffic flow\cite{liu2021community, bai2020adaptive, li2021dynamic, ye2021coupled, han2021dynamic}, ignoring the physical concepts that lead to the generation of observations and the causal relationship between these concepts. Spatio-temporal correlations are considered unstable under the influence of different conditions, and spurious correlations may exist in observations. Causality is necessary when we delve into the generation principle of observation. For example, researchers\cite{ye2019co, deng2021pulse} believe that there is a certain correlation between taxi and bike flow, and that they can boost each other in terms of multi-task learning. As shown in Fig. \ref{fig:1}(b), the flow of taxis and bikes seems to be correlated under normal conditions. Since people's demand for arriving or leaving a region is consistent during rush hours, the tends are similar. However, when it rains (marked in red), the demand for bikes decreases due to weather changes, but the demand for taxis increases, with diametrically opposite trends during the same period. This indicates a spurious correlation between taxis and bike flow under the influence of weather. Second, we believe that the regional attribute has a strong causal relationship with people's travel demand. As shown in Fig. \ref{fig:1}(b) and (c), the area has a strong regional attraction under the influence of the hospital attribute, leading to a large demand for people, so it has obvious rush hours in morning and noon. In addition, this area is chronically congested due to high demand(marked in green). Beijing Financial Street is the main working area with a large number of enterprises, so it has obvious morning and evening rush hours (marked in green). We provide more examples of regional POI elements affecting travel demand in appendix. Finally, the demand for taxis may have an impact on the traffic speed. As shown in Fig.\ref{fig:1}(c). The taxi demand can be inferred from flow. The larger the taxi flow, the more vehicles on the road, and the slower traffic speed (marked in blue). By contrast, High bus demand does not mean a large number of buses on the road, so there is little causal relationship between bus demand and speed.


According to the above analysis, the essential factor affecting multimodal traffic observations is the causal relationship with physical concepts, and excessive attention to the correlation will lead to unstable prediction results. we rethink the generation process of multimodal traffic flow, and explicitly separate the core physical concepts affecting the observation generation into three groups: 1) The attraction factor of the region to people in different time periods. 2) The demand factor (including bikes, taxis and buses) of people choosing different transportation modes under different conditions, and 3) The speed factor affected by the number of vehicles on the road. Our primary task is to disentangle the causal representation of these concepts from conditional information and observations, and further explore their causal relationship.

In this paper, we regard the spatio-temporal multimodal traffic sequence generation process as a Conditional Markov Process, and propose a Causal Conditional Hidden Markov Model (CCHMM). We disentangle the underlying explanatory factors by means of Variational inference, and establish the causal relationship between latent variables by using the Structural Causal Model (SCM)\cite{pearl2009causality, scholkopf2022causality}. Compared with the existing work, instead of building a complex adjacency graph between regions to extract the spatio-temporal correlations in the observation data, we model multimodal traffic flow prediction from a causal perspective. The theoretical innovation in the field of traffic forecasting is as follows: Based on the idea of causality, we model the operation process of multimodal traffic systems from the perspective of the observation generation principle, while the existing methods do not focus on causality in the observation data. We propose a causal graph (shown in Fig. \ref{fig:2}) to describe the operation of multimodal traffic systems, on which we define a joint distribution (shown in Eq. \ref{equ:1}) that describes the principle of observation data generation. Specifically, first, the posterior network infers the disentangled representation of concepts of interest from conditional information and observation data and learns the variational posterior distribution. Then, the prior network models the natural physical laws that existed in the system from the conditional information and learns the prior distribution of the concepts of interest. Third, the causal propagation module mines the causal effects and transforms the exogenous variables inferred from the prior and posterior networks into causal endogenous variables. Finally, The causal endogenous variables are fed into the generator to generate multimodal traffic flow and regarded as the prediction results. The main contributions of this work are as follows:

\begin{itemize}
 \item We analyze the core physical concepts that affect the multimodal traffic flow generation process, disentangle the causal representations of concepts of interest, and further explore their causal relationship.
 \item We reform the previous prediction methods and innovatively propose a Causal Conditional Hidden Markov Model (CCHMM) to predict multimodal traffic flow from the perspective of observation generation principle.
 \item We propose a mutually supervised training method for the prior and posterior to capture physical rules of concepts and enhance the causal identifiability of the model.
 \item Extensive experiments on real-world datasets show that CCHMM comprehensively outperforms state-of-the-art methods for multimodal traffic  flow prediction.

\end{itemize}

\section{Related works}

\textbf{Multimodal traffic flow prediction.} Giving the increasing availability of diverse data sources, most recent studies has focused on the multimodal fusion in traffic flow prediction. Researchers construct models based on multi-task learning framework to forecast traffic flow and speed simultaneously\cite{IOT2021,partc2021}. Ye \cite{ye2019co} et al. decompose spatial traffic flow with a convolutional autoencoder and implement heterogeneous LSTM for predicting traffic flow of three traffic modes simultaneously. Deng \cite{deng2021pulse} et al. learn multi-view representations for single-modal traffic flow and introduce a cross-view self-attention mechanism to capture the co-evolution correlation between different traffic modes. Most of these works implemented Multilayer Perception (MLP) for encoding conditional information(e.g. weather and POI)\, utilized CNN\cite{liang2021fine, cao2021bert} or Graph Convolutional networks (GCN)\cite{graphwavenet2019, han2021dynamic} for capturing spatial features and used RNN for temporal features\cite{ye2021coupled, li2021dynamic,bai2020adaptive}. Finally, the fused features are fed into downstream prediction network. However, these models do not distinguish the features related to different tasks, which make models learn spurious correlations during the training process. The spurious correlations make models difficult to generalize beyond their training distribution.

\textbf{Causal disentangled representation learning.} In representation learning, the observation $x$ is generated by a two-step generative process. First, the latent variable $z$ is sampled from a prior distribution $p(z)$, and then the observation $x$ is sampled from the conditional distribution $p(x|z)$\cite{locatello2019challenging}. Disentangled representation learning aims to learn separable latent variables $z=\left\{z_1,z_2,\dots,z_n\right\}$. Most existing methods rely on the independency assumption of latent variables which is potentially unrealistic\cite{khemakhem2020variational}. In fact, there is generally a complex causal relationship between latent variables\cite{yang2021causalvae}. To address this issue, recent works are proposed to combine SCM \cite{pearl2009causality, scholkopf2022causality} with deep learning models.
CasualVAE \cite{yang2021causalvae} proposes a model with causal layer to transform exogenous factors into causal endogenous ones that correspond to causally related concepts in data. Shen \cite{shen2020disentangled}et al. use a SCM as the prior for bidirectional generative model which can generate data from any desired interventional distributions of the latent factors. Different from above works, our model focused on causal disentangled representation learning on spatial-temporal series. Li \cite{li2021causal} et al. propose a time series disease forecasting method based on HMM. Although this method can disentangle the latent variables that are related to disease, while ignores the casual relationship among factors. In our model, we construct a comprehensive temporal causal graph for conditional information, latent variables and observation data. To the best of our knowledge, our work is the first one that successfully applies the structural causal model to traffic prediction problems.


\section{Methodology}
\label{Methodology}
\subsection{Problem Definition}
\label{Problem Definition}

\begin{figure}
    \centering
    \includegraphics[width=0.4\textwidth]{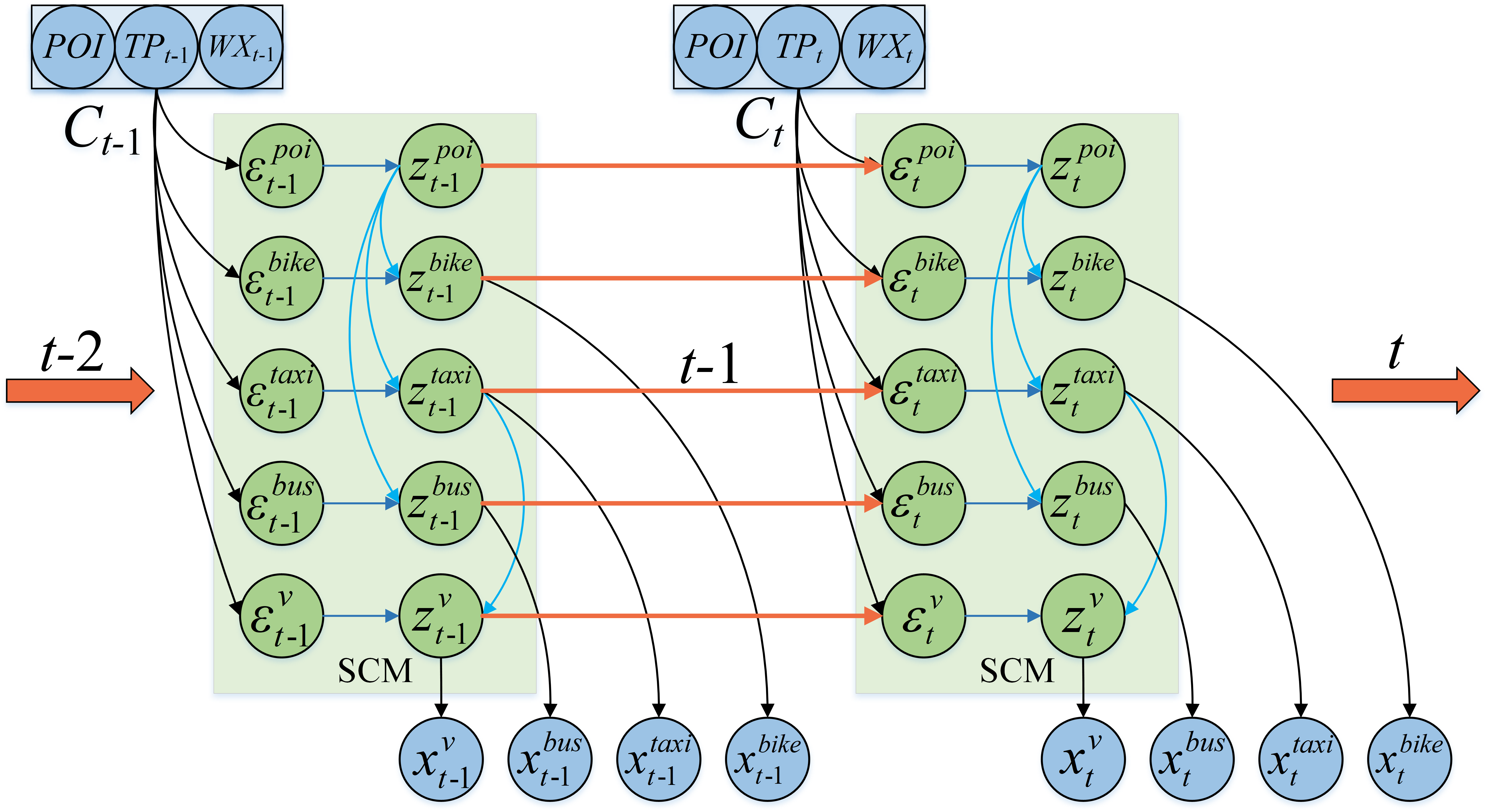}
    \caption{The causal graph of multimodal transportation systems}
    \label{fig:2}
\end{figure}

We define the generation process of multimodal traffic flow as a Conditional Markov Process, illustrated as a Directed Acyclic Graph(DAG), as shown in Fig \ref{fig:2}. For the latent variable inference stage at time step $t$, the conditional information $\mathbf{C}_t$ composed of $POI$, time position $TP_t$ and weather $WX_t$ reflects the current system external status. The conditional information is combined with causal endogenous latent variables $\mathbf{z}_{t-1}$ from the previous time step $t-1$ to extract independent exogenous variables $ \boldsymbol{ \epsilon}_{t}=\left[ \boldsymbol{\epsilon}_{t}^{p o i}, \boldsymbol{\epsilon}_{t}^{{bike }}, \boldsymbol{\epsilon}_{t}^{{taxi }}, \boldsymbol{\epsilon}_{t}^{b u s}, \boldsymbol{\epsilon}_{t}^{{v }}\right] $, which is determined by the system external status and are not affected by observations. Then, the Structural Causal Model (SCM) $\mathbf{z}_i \leftarrow  f(pa(\mathbf{z}_i), \boldsymbol{\epsilon}_i) $\cite{scholkopf2022causality} assigns the generative mechanism of each latent endogenous variable, where $pa(\mathbf{z}_i)$ denotes the set of parent nodes of $\mathbf{z}_i$. It transforms the independent exogenous variables to causal endogenous variables $\mathbf{z}_{t}=\left[ \mathbf{z}_{t}^{p o i}, \mathbf{z}_{t}^{{bike }}, \mathbf{z}_{t}^{{taxi }}, \mathbf{z}_{t}^{{bus }}, \mathbf{z}_{t}^{ {v }}\right]$.
The causal endogenous latent variables $\mathbf{z}_{t}$ are regarded as an approximate representation of a series of concepts of interest, where the elements represent the regional attraction factor, bike demand factor, taxi demand factor, bus demand factor and speed factor at time $t$, respectively.  Since these latent variables evolve as intrinsic drivers for the progression of multimodal traffic observations, the prior distribution of the latent variables has Markov property and is defined as $p( \boldsymbol{ \epsilon}_t, \mathbf{z}_t|\mathbf{z}_{t-1}, \mathbf{C}_t)=p( \boldsymbol{ \epsilon}_t|\mathbf{z}_{t-1}, \mathbf{C}_t) * p(\mathbf{z}_t| \boldsymbol{ \epsilon}_t)$.


For the data generation stage at time step $t$, the exogenous latent variables $\boldsymbol{ \epsilon}_t$ are sampled from the prior distribution $p( \boldsymbol{ \epsilon}_t|\mathbf{z}_{t-1}, \mathbf{C}_t)$, The causal endogenous latent variables $\mathbf{z}_{t}$ are generated using a SCM. Finally, the observations $\mathbf{x}_{t}=\left[\mathbf{x}_{t}^{b i k e}, \mathbf{x}_{t}^{t a x i}, \mathbf{x}_{t}^{b u s}, \mathbf{x}_{t}^{{v }}\right]$ are generated from the conditional distribution $p(\mathbf{x}_t|\mathbf{z}_t)$.

\subsection{A Probabilistic Generative Model for CCHMM}
\label{A Probabilistic Generative Model for CCHMM}
We give the joint distribution definition of the probabilistic generative model of CCHMM and factorize it according to the DAG (Fig. \ref{fig:2}) and Causal Markov Condition\cite{pearl2009causality}:
\begin{equation}\label{equ:1}
    \begin{aligned}
       & p_{\theta}\left(\mathbf{x}_{t<T}, \boldsymbol{ \epsilon}_{t<T}, \mathbf{z}_{t<T} \mid \mathbf{C}_{t<T}\right) \\
        &=\prod_{t=1}^{T-1} p_{\theta}\left( \boldsymbol{ \epsilon}_{t}, \mathbf{z}_{t} \mid \mathbf{z}_{t-1}, \mathbf{C}_{t}\right) * p_{\theta}\left(\mathbf{x}_{t} \mid \mathbf{z}_{t}\right)
    \end{aligned}
\end{equation}
The first term is the prior model, which can be further factored into the generative mechanism of exogenous and endogenous variables based on the causal relationship:
\begin{equation}\label{equ:2}
    \begin{aligned}
        p_{\theta}\left( \boldsymbol{ \epsilon}_{t}, \mathbf{z}_{t} \mid \mathbf{z}_{t-1}, \mathbf{C}_{t}\right)=p_{\theta}\left( \boldsymbol{ \epsilon}_{t} \mid \mathbf{z}_{t-1}, \mathbf{C}_{t}\right)* p_{\theta}\left(\mathbf{z}_{t} \mid \boldsymbol{ \epsilon}_{t}\right)
    \end{aligned}
\end{equation}
The second item is the generative model, which can be further factored into generative models for each modality depending on endogenous variables corresponding to concepts of interest:
\begin{equation}\label{equ:3}
    \begin{aligned}
        p_{\theta}\left(\mathbf{x}_{t} \mid \mathbf{z}_{t}\right)
        &=p_{\theta}\left(\mathbf{x}_{t}^{b i k e} \mid \mathbf{z}_{t}^{b i k e}\right) * p_{\theta}\left(\mathbf{x}_{t}^{t a x i} \mid \mathbf{z}_{t}^{t a x i}\right)  \\
        & * p_{\theta}\left(\mathbf{x}_{t}^{b u s} \mid \mathbf{z}_{t}^{b u s}\right) * p_{\theta}\left(\mathbf{x}_{t}^{v} \mid \mathbf{z}_{t}^{v}\right)
    \end{aligned}
\end{equation}
We apply variational Bayes to learn a tractable distribution $q_{\phi}$ to approximate the true posterior $p_{\theta}$, defined as follows:
\begin{equation}\label{equ:4}
    \begin{aligned}
    q_{\phi}\left( \boldsymbol{ \epsilon}_{t<T}, \mathbf{z}_{t<T} \mid \mathbf{x}_{t<T}, \mathbf{C}_{t<T}\right) &=\prod_{t=1}^{T-1} q_{\phi}\left( \boldsymbol{ \epsilon}_{t} \mid \mathbf{z}_{t-1}, \mathbf{x}_{t}, \mathbf{C}_{t}\right) \\
    &* q_{\phi}\left(\mathbf{z}_{t} \mid \boldsymbol{ \epsilon}_{t}\right)
    \end{aligned}
\end{equation}

\subsection{Causal Conditional Hidden Markov Model}
\label{Causal Conditional Hidden Markov Model}
\begin{figure*}
  \centering
  \includegraphics[width=0.90\textwidth]{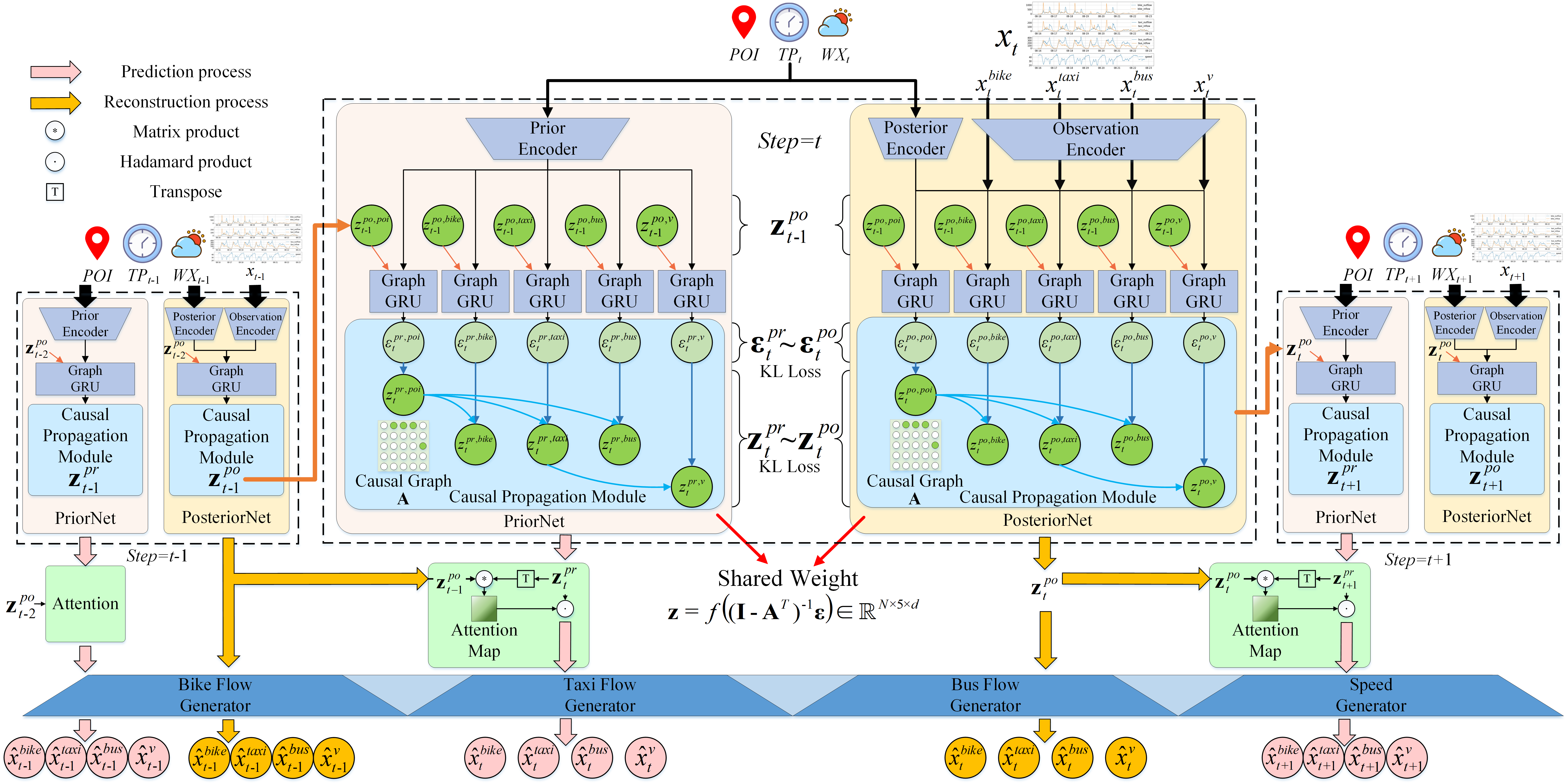}
  \caption{The architecture of CCHMM.}
  \label{fig:3}
\end{figure*}

To model Causal Conditional Hidden Markov Model based on the above probabilistic generative model, as shown in Fig. \ref{fig:3}, our main tasks are as follows: (1) In the latent variable inference stage, a deep neural network is used to fit the prior distributions $p_{\theta}\left( \boldsymbol{ \epsilon}_{t}, \mathbf{z}_{t} \mid \mathbf{z}_{t-1}, \mathbf{C}_{t}\right)$ and posterior distributions $q_{\phi}\left( \boldsymbol{ \epsilon}_{t}, \mathbf{z}_{t} \mid \mathbf{z}_{t-1}, \mathbf{x}_{t}, \mathbf{C}_{t}\right)$ of latent variables to disentangle the causal representations of concepts affecting the generation of multimodal traffic observations. (2) A causal propagation module is proposed to mine the causal relationship between endogenous latent variables through a trainable causal graph, and propagate the causal effect according to the causal order. (3) In the observation data generation stage, the generator is established to approximate the conditional generation distribution $p_\theta(\mathbf{x}_t|\mathbf{z}_t)$. We utilize learnable variational distributions to approximate the true data distribution, with the aim of disentangling causal representations of physical concepts using variational inference. Compared to traditional VAE, we explicitly endow latent variables with real semantic information (i.e., causal representations of physical concepts).

\subsection{Posterior Network}
\label{Posterior Network}
We use conditional information and observations to build a PosteriorNet, whose purpose is to approximate the true posterior distribution of latent variables by learning a variational posterior distribution $q_{\phi}\left( \boldsymbol{ \epsilon}_{t}, \mathbf{z}_{t} \mid \mathbf{z}_{t-1}, \mathbf{x}_{t}, \mathbf{C}_{t}\right)$ using neural networks. As shown in the yellow part of Fig. \ref{fig:3}, it consists of Graph Gated Recurrent Unit (GraphGRU) and Causal Propagation Module.

\paragraph{GraphGRU}

The progression of multimodal traffic flow has Markov property, and the evolution of latent variables is the intrinsic drivie for the spatio-temporal dependencies of multimodal traffic observations. Therefore, we use the GraphGRU to model the evolution process of system status, capturing the spatio-temporal dependencies into the exogenous latent variables. We build parameter-independent GraphGRU to learn mode-specific patterns for each traffic modes, defined as follows:
\begin{equation}\label{equ:5}
    \begin{aligned}
        \mathbf{s}_{t}^{po,i} &= \operatorname{FC}(\mathbf{C}_t||\mathbf{x}_t^i)  \\
        {\mathbf{r}_{t}^{po,i}} & = \sigma ({\mathbf{W}_{r}^i}{{\star }_{G}}({\mathbf{s}_{t}^{po,i}}||{\mathbf{z}_{t-1}^{po,i}})+{\mathbf{b}_{r}^i})  \\
        {{\mathbf{u}}_{t}^{po,i}} & = \sigma ({\mathbf{W }_{u}^i}{{\star }_{G}}({\mathbf{s}_{t}^{po,i}}||{\mathbf{z}_{t-1}^{po,i}})+{\mathbf{b}_{u}^i}) \\
        {\mathbf{\tilde{h}}_{t}^{po,i}} & = \operatorname{tanh} ({\mathbf{W }_{h}^i}{{\star }_{G}}({\mathbf{s}_{t}^{po,i}}||(\mathbf{r}_{t}^{po,i}\odot {\mathbf{z}_{t-1}^{po,i}}))+{\mathbf{b}_{h}^i}) \\
        {\boldsymbol{ \epsilon}_{t}^{po,i}} & = {{\mathbf{u}}_{t}^{po,i}}\odot {\mathbf{z}_{t-1}^{po,i}}+(1-{\mathbf{u}}_{t}^{po,i})\odot {\mathbf{\tilde{h}}_{t}^{po,i}}
    \end{aligned}
\end{equation}
where $i \in \left \{poi, bike, taxi, bus, v \right \}$ denotes physical concept of interest, $||$ denotes concatenate operation, $\sigma$ denotes sigmoid funtion, $\mathbf{C}_t \in \mathbb{R}^{N\times c_c} $ is conditional information, $\mathbf{x}_t^i \in \mathbb{R}^{N\times c_i}$ is the observation of the $i$-th mode, $c_i$ is the number of traffic flow channels of the $i$-th mode, ${\mathbf{z}_{t-1}^{po,i}} \in \mathbb{R}^{N\times d}$ is the posterior endogenous latent variable at $t-1$, $\boldsymbol{ \epsilon}_{t}^{po,i} \in \mathbb{R}^{N\times d}$ is posterior exogenous latent variable at $t$, and $\mathbf{W}, \mathbf{b}$ are the parameters of graph convolution. The graph convolution defined by $\mathbf{W}\star_{G}(\mathbf{X}) + \mathbf{b} = (\mathbf{I}+\mathbf{D}^{-1/2}\mathbf{G}\mathbf{D}^{-1/2})\mathbf{X}\mathbf{W}+\mathbf{b} $, where $\mathbf{G} \in \mathbb{R}^{N \times N}$ is the distance adjacency matrix of regions, ${{\mathbf{D}}_{ii}}=\sum\nolimits_{j}{{{\mathbf{G}}_{ij}}}$ and $N$ is the number of regions. Then, we calculate the mean and log-variance of $ \boldsymbol{ \epsilon}_{t}^{po}$ by using separate fully connected layers for each traffic modes to obtain the posterior distribution of exogenous latent variables $q_{\phi}\left(\boldsymbol{ \epsilon}_{t} \mid \mathbf{z}_{t-1}, \mathbf{x}_{t}, \mathbf{C}_{t}\right)$.


\paragraph{Causal Propagation Module}

Concepts that affect the generation of observations are naturally causally related. Therefore, endogenous latent variables, as semantic representations of concepts, also have causal relationships. We propose a causal propagation module to transform independent exogenous variables into causal endogenous variables and leverage a learnable causal graph to mine their causal relationships.

The linear SCM is defined as $\mathbf{z}={{\mathbf{A}}^{T}}\mathbf{z}+\boldsymbol{\epsilon }={{(\mathbf{I-}{{\mathbf{A}}^{T}})}^{-1}}\boldsymbol{\epsilon }$. We add parameter-independent nonlinear transformations for each traffic modes to improve the representation ability. In this paper, the causal propagation module is defined as:
\begin{equation}\label{equ:6}
    \begin{aligned}
    \tilde{\mathbf{A}} &=\operatorname{ReLU}\left(\tanh \left(\alpha \mathbf{W}_{\mathbf{A}}\right)\right) \in \mathbb{R}^{5 \times 5} \\
    \mathbf{h}_{t}^{p o} &=\left(\mathbf{I}-\tilde{\mathbf{A}}^{T}\right)^{-1} \boldsymbol{\epsilon}_{t}^{p o} \in \mathbb{R}^{N  \times 5 \times d} \\
    \mathbf{z}_{t}^{p o, i} &=f_{i}\left(\left[\mathbf{h}_{t}^{p o}\right]_{\left[:,i,:\right]}\right)=\mathrm{FC}_{i}\left(\tanh \left(\mathrm{FC}_{i}\left(\left[\mathbf{h}_{t}^{p o}\right]_{\left[:,i,:\right]}\right)\right)\right)
    \end{aligned}
\end{equation}
where $ \mathbf{W}_{\mathbf{A}} \in \mathbb{R}^{5 \times 5} $ denotes a learnable parameter, $\alpha$ is a hyper-parameter for controlling the saturation rate of the activation function. ReLU regularizes the parameter matrix to ensure sparsity and non-negativity. $\tilde{\mathbf{A}}$ is the causal graph of endogenous latent variables, where $\tilde{\mathbf{A}}_{ij}$ represents the causal effect of the parent variable $\mathbf{z}_i$ on the child variable $\mathbf{z}_j$. Therefore, when the graph nodes are sorted in topological order, the matrix $\tilde{\mathbf{A}}$ is strictly upper triangular. Then, we calculate the mean and log-variance by using separate fully connected layers for each traffic modes.


\subsection{Prior Network}
\label{Prior Network}
Previous unsupervised disentangled representation learning based on VAE regularizes the posterior of the latent variables with a standard Multivariate Gaussian prior, which greatly limits the expression ability of the model. Unsupervised disentangled representation learning can not guarantee the model identifiability due to the lack of inductive bias \cite{locatello2019challenging}. To improve the identifiability of the model, we build a PriorNet based on conditional information, which aims to model the physical rules of the concepts of interest that naturally exist in the system, and use a learnable prior distribution to approximate this rules. As shown in the pink part of Fig. \ref{fig:3}, the PirorNet is similar in structure to the PosteriorNet, which is composed of GraphGRU and causal propagation module.

\paragraph{GraphGRU}
The PriorNet only inputs the conditional information of the current system, calculates prior exogenous latent variables $ \boldsymbol{ \epsilon}_{t}^{pr}$ according to Eq.\ref{equ:5}, and then obtains the prior distribution of exogenous latent variables $p_{\theta}\left(\boldsymbol{ \epsilon}_{t} \mid \mathbf{z}_{t-1}, \mathbf{C}_{t}\right)$ by calculating the mean and log-variance.

\paragraph{Causal Propagation Module}
The PriorNet and the PosteriorNet share a causal propagation module. We argue that causality is a stable natural phenomenon that does not change with time or space, thus globally sharing a causal graph and nonlinear transformation. We calculate the prior endogenous latent variables $\mathbf{z}_{t}^{p r}$ according to Eq.\ref{equ:6}, and then obtain the prior distribution of the endogenous latent variables $p_{\theta}\left(\mathbf{z}_{t} \mid \boldsymbol{\epsilon}_t\right)$ by calculating the mean and log-variance.

\subsection{Generator}
\label{Generator}
We build the generator using two fully connected layers to parameterize the conditional distribution of the generative models $p_{\theta}\left(\mathbf{x}_{t} \mid \mathbf{z}_{t}\right)$ defined in Eq.\ref{equ:3}. As shown in Figure 3, a generator is globally shared. The generator is shared globally as shown in Fig. \ref{fig:3}. The results of generative models have different meanings depending on the type of $\mathbf{z}$.
\paragraph{Reconstruction}
As shown by the yellow arrow in Fig. \ref{fig:3}, the PosteriorNet takes the current observations as part of the input. So when generating data using the posterior endogenous latent variables $\mathbf{z}_{t}^{p o}$, the output is the reconstruction result, represented as $\mathbf{\hat{x}}_{t}^{i,rec}=\operatorname{Generator}_i (\mathbf{z}_{t}^{po,i})$, $i \in \left \{bike, taxi, bus, v \right \}$.

\paragraph{Prediction}
The PriorNet only utilizes the current conditional information to fit the prior distribution and does not involve current observations. Therefore, when generating data using the prior endogenous latent variables $\mathbf{z}_{t}^{p r}$, the output is the prediction result. Based on the Markov property of sequence generation, we leverage a simple attention mechanism to weight the current prior endogenous latent variables with the previous posterior, which can further improve the effect of prediction. The attention mechanism is defined as:
\begin{equation}\label{equ:7}
    \begin{aligned}
    \mathbf{\tilde{z}}_{t}^{pr}=\operatorname{softmax}\left( \mathbf{z}_{t-1}^{po}{{\mathbf{W}}_{att}}{{\left( \mathbf{z}_{t}^{pr} \right)}^{T}} \right)\mathbf{z}_{t}^{pr} \in \mathbb{R}^{N \times 5 \times d}
    \end{aligned}
\end{equation}
where $\mathbf{W}_{att} \in \mathbb{R}^{ d \times d}$ is the learnable parameter. Then $\mathbf{\tilde{z}}_{t}^{pr}$ is fed into the generator to obtain the prediction result, represented as $\mathbf{\hat{x}}_{t}^{i,pred}=\operatorname{Generator}_i (\mathbf{\tilde z}_{t}^{pr,i})$, $i \in \left \{bike, taxi, bus, v \right \}$.

\subsection{Learning Strategy}
\label{Learning Strategy}
We propose a mutually supervised training method for the PriorNet and PosteriorNet, which benefits the model to approximate the physical rules of concepts of interest, while helping the to identifiably disentangle causal representations. Based on variational inference, we use a neural network to learn a tractable distribution $q_{\phi}$ to approximate the true posterior distribution $p_{\theta}$. Given a dataset $\mathcal{D}$, the Evidence Lower Bound (ELBO) of CCHMM is as follows:
\begin{small}
\begin{equation}\label{equ:8}
    \begin{aligned}
        \mathcal{L}_{ELBO}&={{\mathbb{E}}_\mathcal{D}}\left[ {{\mathbb{E}}_{{{q}_{\phi }}}}\left[ \log \left( \frac{{{p}_{\theta }}({{\mathbf{x}}_{t<T}},{{\boldsymbol{\epsilon }}_{t<T}},{{\mathbf{z}}_{t<T}}|\mathbf{C}_{t<T})}{{{q}_{\phi }}({{\boldsymbol{\epsilon }}_{t<T}},{{\mathbf{z}}_{t<T}}|{{\mathbf{x}}_{t<T}},\mathbf{C}_{t<T})} \right) \right] \right] \\
        &=   {{\mathbb{E}}_\mathcal{D}}\left[ \sum\limits_{t=1}^{T-1}\mathcal{L}_t^{q_\phi, p_\theta } \right]\\
        \mathcal{L}_t^{q_\phi, p_\theta }&= {{{\mathbb{E}}_{{{q}_{\phi }}\left( {{\boldsymbol{\epsilon }}_{t}},{{\mathbf{z}}_{t}}|{{\mathbf{z}}_{t-1}},{{\mathbf{x}}_{t}},{{\mathbf{C}}_{t}} \right)}}}\left[ \log \left( {{p}_{\theta }}({{\mathbf{x}}_{t}}|{{\mathbf{z}}_{t}}) \right) \right]\\
        &-{{D}_{KL}}\left[ {{q}_{\phi }}\left( {{\boldsymbol{\epsilon }}_{t}},{{\mathbf{z}}_{t}}|{{\mathbf{z}}_{t-1}},{{\mathbf{x}}_{t}},{{\mathbf{C}}_{t}} \right)||{{p}_{\theta }}\left( {{\boldsymbol{\epsilon }}_{t}},{{\mathbf{z}}_{t}}|{{\mathbf{z}}_{t-1}},\mathbf{C}_{t} \right) \right]
    \end{aligned}
\end{equation}
\end{small}
We rewrite Eq. \ref{equ:6} as ${{\mathbf{z}}_{t}}={{\varphi }_{\mathbf{w}}}({{\mathbf{\epsilon }}_{t}})$, where $\mathbf{w}$ is the parameter of the causal propagation module and ${\varphi }_{\mathbf{w}}$ is invertible. Therefore, we reformulate the of the prior and posterior distributions with the Dirac delta function $\delta (\cdot )$, represented as follows:
\begin{small}
\begin{equation}\label{equ:9}
    \begin{aligned}
         {{q}_{\phi }}\left( {\boldsymbol{\epsilon}_{t}},{{\mathbf{z}}_{t}}|{{\mathbf{z}}_{t-1}},{{\mathbf{x}}_{t}},{{\mathbf{C}}_{t}} \right)&={{q}_{\phi }}\left( {\boldsymbol{\epsilon}_{t}}|{{\mathbf{z}}_{t-1,}}{{\mathbf{x}}_{t}},{{\mathbf{C}}_{t}} \right)\delta \left( {{\mathbf{z}}_{t}}={{\varphi }_{\mathbf{w}}}({\boldsymbol{\epsilon}_{t}}) \right)\\
         &={{q}_{\phi }}\left( {{\mathbf{z}}_{t}}|{{\mathbf{z}}_{t-1,}}{{\mathbf{x}}_{t}},{{\mathbf{C}}_{t}} \right)\delta \left( {\boldsymbol{\epsilon}_{t}}=\varphi_{\mathbf{w}}^{\text{-}1}({{\mathbf{z}}_{t}}) \right) \\
        {{p}_{\theta }}\left( {\boldsymbol{\epsilon}_{t}},{{\mathbf{z}}_{t}}|{{\mathbf{z}}_{t-1}},{{\mathbf{C}}_{t}} \right)&={{p}_{\theta }}\left( {\boldsymbol{\epsilon}_{t}}|{{\mathbf{z}}_{t-1}},{{\mathbf{C}}_{t}} \right)\delta \left( {{\mathbf{z}}_{t}}={{\varphi }_{\mathbf{w}}}({\boldsymbol{\epsilon}_{t}}) \right)\\
        &={{p}_{\theta }}\left( {{\mathbf{z}}_{t}}|{{\mathbf{z}}_{t-1}},{{\mathbf{C}}_{t}} \right)\delta \left( {\boldsymbol{\epsilon}_{t}}=\varphi _{\mathbf{w}}^{\text{-}1}({{\mathbf{z}}_{t}}) \right)
    \end{aligned}
\end{equation}
\end{small}
We substitute the prior and the posterior distributions in Eq. \ref{equ:9} and reformulate $\mathcal{L}_t^{q_\phi, p_\theta }$ as:
\begin{equation}\label{equ:10}
    \begin{aligned}
        \mathcal{L}_t^{q_\phi, p_\theta }= &{{{\mathbb{E}}_{{{q}_{\phi }}\left( {{\mathbf{z}}_{t}}|{{\mathbf{z}}_{t-1}},{{\mathbf{x}}_{t}},{{\mathbf{C}}_{t}} \right)}}}\left[ \log \left( {{p}_{\theta }}({{\mathbf{x}}_{t}}|{{\mathbf{z}}_{t}}) \right) \right]\\
        &-{{D}_{KL}}\left[ {{q}_{\phi }}\left( {{\boldsymbol{\epsilon}}_{t}}|{{\mathbf{z}}_{t-1}},{{\mathbf{x}}_{t}},{{\mathbf{C}}_{t}} \right)||{{p}_{\theta }}\left( {{\boldsymbol{\epsilon}}_{t}}|{{\mathbf{z}}_{t-1}},\mathbf{C}_{t} \right) \right] \\
        &-{{D}_{KL}}\left[ {{q}_{\phi }}\left( {{\mathbf{z}}_{t}}|{{\mathbf{z}}_{t-1}},{{\mathbf{x}}_{t}},{{\mathbf{C}}_{t}} \right)||{{p}_{\theta }}\left( {{\mathbf{z}}_{t}}|{{\mathbf{z}}_{t-1}},\mathbf{C}_{t} \right) \right]
    \end{aligned}
\end{equation}
where, the first term is the reconstruction loss, and the last two terms are the KL divergence of the exogenous and endogenous latent variables, respectively.

Since the causal graph has the property of being acyclic, it is necessary to increase the acyclic constraint \cite{yu2019dag} of $\tilde{\mathbf{A}}$, expressed as $h(\tilde{\mathbf{A}}) = \operatorname{tr}\left[(I+\tilde{\mathbf{A}} \circ \tilde{\mathbf{A}})^{n}\right]-n$. In addition, we use L2-norm as the predicted loss, defined as $\mathcal{L}^{{pred }}=\mathbb{E}_{\mathcal{D}}\left[\sum_{t=1}^{T-1}\left\|\hat{\mathbf{x}}_{t}^{pred }-\mathbf{x}_{t}\right\|_{2}^{2}\right]$. In summary, the total loss function of CCHMM is defined as follows:
\begin{equation}\label{equ:11}
    \begin{aligned}
    \mathcal{L}=-\mathcal{L}_{ELBO}+\mathcal{L}_{pred}+\lambda h(\tilde{\mathbf{A}})
    \end{aligned}
\end{equation}
where $\lambda$ is hyper-parameter for controlling the loss balance.

\section{Experiments}
\label{Experiments}

We evaluate the performance of our model on real world traffic datasets and compare with some recent compelling methods for traffic flow prediction\footnote{\url{https://github.com/EternityZY/CCHMM}}. Further, a comprehensive ablation study shows the effectiveness of each component of our model.

\begin{table*}
\caption{Performance comparison with other models.}
\label{tab3}
\resizebox{\hsize}{!}{
\begin{tabular}{c|ccc|ccc|ccc|ccc}
\hline
\multirow{2}{*}{Models} & \multicolumn{3}{c|}{Bike}                              & \multicolumn{3}{c|}{Taxi}                              & \multicolumn{3}{c|}{Bus}                                & \multicolumn{3}{c}{Speed}                             \\ \cline{2-13}
                        & MAE             & RMSE            & MAPE               & MAE             & RMSE            & MAPE               & MAE             & RMSE             & MAPE               & MAE             & RMSE            & MAPE              \\ \hline
HMM                     & 6.3239          & 13.1028         & 24.8122\%          & 5.0146          & 8.6146          & 26.8858\%          & 6.7976          & 13.0332          & 21.2048\%          & 1.3766          & 2.2260          & 4.3131\%          \\
HGCN                    & 5.6612          & 10.5526         & 23.2818\%          & 4.8627          & 8.5967          & 25.4933\%          & 7.5434          & 14.7726          & 22.4939\%          & 1.9353          & 2.8477          & 5.9296\%          \\
CCRNN                   & 5.3530          & 11.34111        & 21.1122\%          & 4.7581          & 8.7107          & 24.7395\%          & 6.6719          & 13.4522          & 20.2636\%          & 1.5560          & 2.5605          & 4.8916\%          \\
DMSTGCN                 & 5.2675          & 9.9759          & 21.5044\%          & 4.5879          & 7.9499          & 24.2042\%          & 6.3610          & 12.1108          & 19.9572\%          & 1.4072          & 2.2511          & 4.3595\%          \\
AGCRN                   & 5.0185          & 9.3577          & 20.3816\%          & 4.5611          & 7.8992          & 23.9883\%          & 6.5580          & 12.5084          & 19.9864\%          & 1.3678          & 2.1587          & 4.2762\%          \\
DGCRN                   & 4.9378          & 9.1436          & 20.3287\%          & 4.5360          & 7.8984          & 23.9846\%          & 6.4283          & 12.2228          & 19.6494\%          & 1.4154          & 2.2878          & 4.4090\%          \\
\textbf{CCHMM(our)}     & \textbf{4.6418} & \textbf{8.5213} & \textbf{19.4286\%} & \textbf{4.4150} & \textbf{7.6262} & \textbf{23.5661\%} & \textbf{6.2450} & \textbf{11.8570} & \textbf{19.2033\%} & \textbf{1.2433} & \textbf{1.9943} & \textbf{3.8588\%} \\ \hline
\end{tabular}}
\end{table*}

\subsection{Dataset}
\label{Dataset}
\textbf{XC-Trans}:The XC-Trans dataset contains order records of three traffic modes(bike, bus and taxi) from June 1st 2021 to December 31th 2021 in Xicheng District, Beijing. The researched region is split into 175 non-overlapping subregions. We statistics the inflow and outflow for each traffic modes in all of the subregions.

\textbf{XC-Speed}:The XC-speed dataset contains speed records of main roads from June 1st 2021 to December 31th 2021 in Xicheng District, Beijing. We use the average speed of road segments within each region to represent the regional speed in every 30 minutes.

Besides, corresponding meteorological information, time position and POI data are collected as conditional information. We split this dataset with a 30-minute interval to obtain 11753 samples. we use three-hour historical data to predict the next 30-minute data. 60\% of the data is used for training, 20\% is used for validating and the rest is used for testing.


\subsection{Experimental settings}
We compare our framework with the following methods.
1)\textbf{HMM}\cite{li2021causal}: It uses multimodal information to achieve robust prediction of irreversible disease at an early stage.
2) \textbf{AGCRN}\cite{bai2020adaptive}: It employs an adaptive graph and integrates GRU with graph convolutions.
3) \textbf{CCRNN}\cite{ye2021coupled}: It employs coupled layer-wise graph convolution layer to capture the multi-level spatial dependence and temporal dynamics simultaneously.
4) \textbf{DGCRN}\cite{li2021dynamic}: It generates a dynamic graph by combining the predefined adjacency matrix and input features.
5) \textbf{HGCN}\cite{guo2021hierarchical}: It constructs road graph and region graph from micro and macro view respectively.
6) \textbf{DMSTGCN}\cite{han2021dynamic}: It designs an adaptive graph construction method to learn the time-specific spatial dependencies of road segments.

\subsection{Overall Comparison}
\label{Overall Comparison}

We evaluate the performance of methods with Mean Absolute Error (MAE), Root Mean Square Error (RMSE) and Mean Absolute Percentage Error(MAPE). Table \ref{tab3} presents the overall prediction performances which are the averaged results over three independent experiments. There is no method compatible with all traffic modes except us.

The baseline models focus on adaptively or dynamically generating graph structures, while our model pay more attention to modeling the causality between latent semantic variables in traffic system. Due to the lack of modeling spatial dependency and causality, the HMM model shows the worst performance. The model based on dynamic graph(e.g. DGCRN) perform better than models based on adaptive graph(e.g. AGCRN). Besides, it can be observed that our model outperforms baseline models consistently and overwhelmingly. Especially in speed prediction, our CCHMM brings about 10\% improvements to the best results in all metrics due to the causality of speed factor is more clear

\subsection{Ablation Study}\label{Ablation Study}

\begin{table*}[]
\caption{Results of ablation study.}
\label{tab4}
\resizebox{\hsize}{!}{
\begin{tabular}{ccccc|ccc|ccc|ccc}
\hline
\multirow{2}{*}{Category}                                                             & \multirow{2}{*}{Models} & \multicolumn{3}{c|}{Bike}                                                                 & \multicolumn{3}{c|}{Taxi}                              & \multicolumn{3}{c|}{Bus}                                & \multicolumn{3}{c}{Speed}                             \\ \cline{3-14}
                                                                                      &                         & MAE                        & RMSE                        & MAPE                           & MAE             & RMSE            & MAPE               & MAE             & RMSE             & MAPE               & MAE             & RMSE            & MAPE              \\ \hline
\multirow{4}{*}{PriorNet}                                                             & w/o GRU                 & 10.4177                    & 22.7940                     & 39.8982\%                      & 9.4037          & 17.6233         & 47.0230\%          & 10.7402         & 21.5010          & 28.9726\%          & 1.9951          & 3.0643          & 6.4062\%          \\
                                                                                      & w/o GCN                 & 6.3110                     & 12.6767                     & 24.8612\%                      & 5.2715          & 9.3825          & 27.1115\%          & 6.7583          & 12.9373          & 20.9419\%          & 1.4340          & 2.3319          & 4.5195\%          \\
                                                                                      & w/o Cond                & 5.7693                     & 11.1578                     & 22.9648\%                      & 5.5484          & 9.9345          & 28.4668\%          & 7.1146          & 13.8885          & 21.5514\%          & 1.4798          & 2.3978          & 4.6165\%          \\
                                                                                      & w/o Prior               & \multicolumn{1}{l}{5.4904} & \multicolumn{1}{l}{10.6453} & \multicolumn{1}{l|}{21.8347\%} & 5.0789          & 9.0566          & 26.1980\%          & 6.8861          & 13.2697          & 21.1046\%          & 1.4735          & 2.4378          & 4.5712\%          \\ \hline
\multirow{3}{*}{\begin{tabular}[c]{@{}c@{}}Causal Propagation \\ Module\end{tabular}} & Entangle                & 5.5317                     & 10.4375                     & 22.5408\%                      & 5.1505          & 9.0548          & 26.9062\%          & 6.9659          & 13.1916          & 21.7755\%          & 1.5514          & 2.5234          & 4.8609\%          \\
                                                                                      & w/o SCM                 & 5.1907                     & 9.5034                      & 21.6663\%                      & 4.9019          & 8.4785          & 26.3348\%          & 6.6320          & 12.6021          & 20.4787\%          & 1.4398          & 2.3503          & 4.4778\%          \\
                                                                                      & w/o non-linear           & 4.9508                     & 8.7033                      & 20.9008\%                      & 4.5376          & 7.7674          & 24.2565\%          & 6.5281          & 12.4300          & 20.1166\%          & 1.3443          & 2.1769          & 4.1931\%          \\ \hline
Our                                                                                   & CCHMM                   & \textbf{4.6418}            & \textbf{8.5213}             & \textbf{19.4286\%}             & \textbf{4.4150} & \textbf{7.6262} & \textbf{23.5661\%} & \textbf{6.2450} & \textbf{11.8570} & \textbf{19.2033\%} & \textbf{1.2433} & \textbf{1.9943} & \textbf{3.8588\%} \\ \hline
\end{tabular}}
\end{table*}

To evaluate the effectiveness of key components, we conduct comprehensive ablation experiments. For PriorNet, we design four variants: 1)w/o GRU: This variant replaces GraphGRU with GCN. The prior of latent variables is only generated from conditional information, which means discarding long-term temporal dependencies. 2)w/o GCN: This variant removes GCN in GraphGRU, which means discarding spatial dependencies. 3) w/o Cond: This variant removes conditional information. Note that we consider $\epsilon$ as exogenous variables which are relevant to conditional information. Removing conditional information is equivalent to removing PriorNet and generating latent variables directly from observation data in PostierNet. 4) w/o Prior: This variant removes the PriorNet but retains conditional information. Different from variant 3, the latent variables are generated from both conditional information and observation with SCM.
For the causal propagation module, we design three variants: 5) Entangle: There is only one latent variable in this variant. 6) w/o SCM: This variant removes SCM, which means the latent variables are directly generated from conditional information and observation. 7) w/o non-linear: This variant replaces the non-linear transformation with linear transformation in SCM.
Note that except for variant 3 and variant 4 that use additional FC layers for prediction, other networks use generator to obtain prediction results.

\begin{figure}[h]
\centering
  \includegraphics[width=0.45\textwidth]{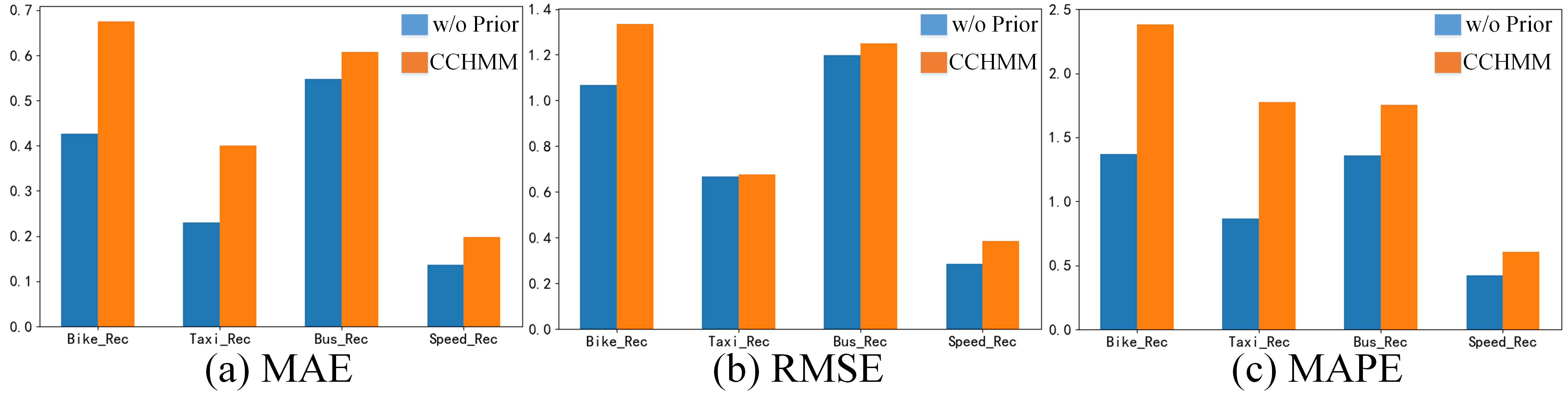}
  \caption{Comparison of Reconstruction performance of w/o Prior and CCHMM.}
  \label{fig9}
\end{figure}

The performance of ablation experiments is shown in Table \ref{tab4}. We can find that variant 1 and 2 perform worst of all due to the lack of spatial and temporal dependencies. The performance of variant 3 shows the necessity of conditional information. In fact, the exogenous variables only affect the system, but are not constrained by the system. It means that we can only determine them by conditional information. Eventually, the model without conditional information degenerates into an ordinary sequence disentangled representation learning model. In variant 4, we drop the PriorNet. The role of the PriorNet is to obtain the stable rules of physical concepts, while the PosteriorNet is designed for obtaining disentangled representations from observation data and conditional information. Posterior collapse may occur in the absence of prior supervision, resulting in failure to obtain a stable and effective causal representation. An evidence is shown in Fig. \ref{fig9}, the reconstruction loss of variant 4 is generally lower than our CCHMM, which means that the model prefers to learn a representation for reconstruction rather than disentangling.

For causal propagation module, the model with disentangle latent variables perform better than the entangle one, which means that VAE-based structures decouple the latent variables to some extent. Since there is no restriction of causal structure, it suffers from spurious correlation. The most obvious consequence is that the speed prediction performance is reduced by 15\%. Besides, the performance of variant 7 linear model is insufficient to express causal relationships in complex scenarios.

\begin{figure}[ht]
	
	\begin{minipage}{0.25\linewidth}
		\vspace{-1pt}
		\centerline{\includegraphics[width=\textwidth]{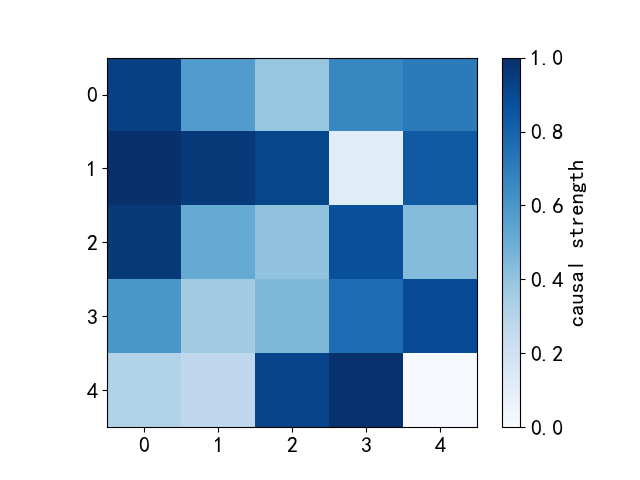}}
		\centerline{(a) w/o GRU}
	\end{minipage}
	\begin{minipage}{0.25\linewidth}
		\vspace{-1pt}
		\centerline{\includegraphics[width=\textwidth]{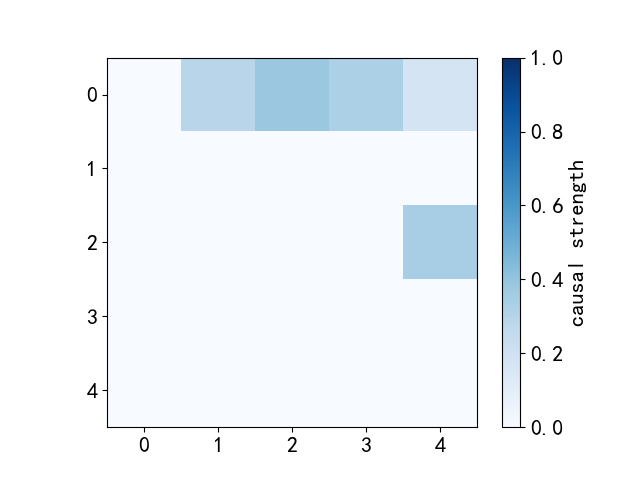}}
	
		\centerline{(b) w/o GCN}
	\end{minipage}
	\begin{minipage}{0.25\linewidth}
		\vspace{-1pt}
		\centerline{\includegraphics[width=\textwidth]{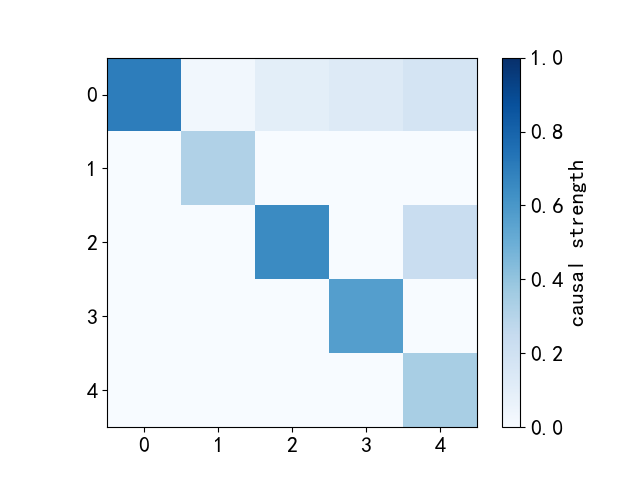}}
	
		\centerline{(c) w/o Prior}
	\end{minipage}
	\centering
	\begin{minipage}{0.25\linewidth}
		\vspace{-1pt}
		\centerline{\includegraphics[width=\textwidth]{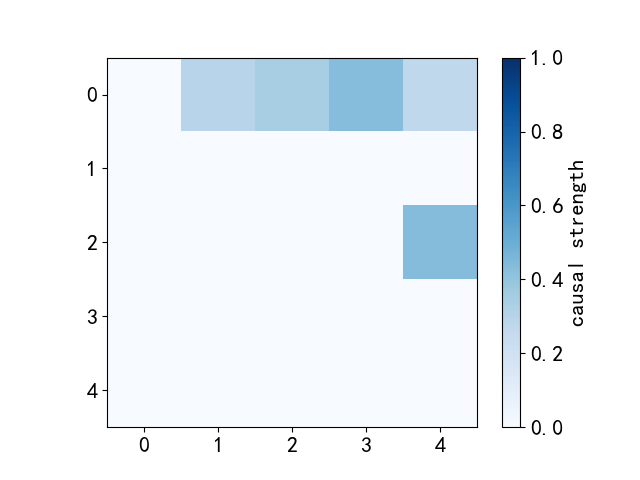}}
	
		\centerline{(d) w/o Nonlinear}
	\end{minipage}
	\begin{minipage}{0.25\linewidth}
		\vspace{-1pt}
		\centerline{\includegraphics[width=\textwidth]{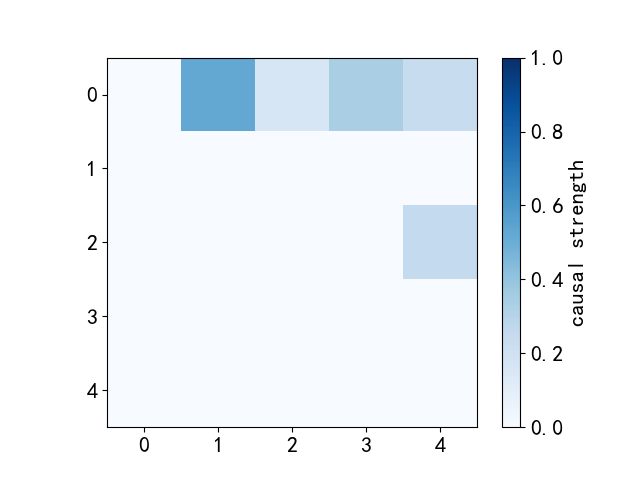}}
	
		\centerline{(e) CCHMM}
	\end{minipage}

	\caption{Visual comparisons of learned causal graphs}
	\label{fig10}
\end{figure}
In addition, for each model with causal propagation module, we initialize the causal graph as an upper triangular matrix subject to standard normal distribution. As shown in Fig. \ref{fig10}, it can be observed that the variant 1 failed to learn a stable causal relationship. The model without GCN and the one without non-linear transformation learnt a causal graph similar to our CCHMM. Particularly, the model without PriorNet learnt an causal graph with large diagonal elements. It means that the model failed to learn representations of physical concepts that conform to causality.

\section{Conclusion}
\label{Conclusion}
In this paper, we analyze the core physical concepts affecting the generation of multimodal traffic flow and disentangle the concepts of interest into three groups: regional attraction factor, the transportation demand factor and traffic speed factor. We infer causal representations of these concepts from conditional information and observations of the current system based on variational inference and structural causal model, and mine their causal relationships by using learnable causal graphs. For the data generation stage, we feed the prior causal representation into the generator to generate predictions. Extensive experiments show that all metrics of CCHMM are optimal, which reveal that it is crucial to introduce causal theory into spatio-temporal sequence analysis. In future work we further explore causal discovery and refine causal relationships in multimodal traffic systems.

\bibliography{reference}

\begin{thebibliography}{27}
\providecommand{\natexlab}[1]{#1}

\bibitem[{Bai et~al.(2020)Bai, Yao, Li, Wang, and Wang}]{bai2020adaptive}
Bai, L.; Yao, L.; Li, C.; Wang, X.; and Wang, C. 2020.
\newblock Adaptive graph convolutional recurrent network for traffic
  forecasting.
\newblock \emph{Advances in Neural Information Processing Systems}, 33:
  17804--17815.

\bibitem[{Cao et~al.(2021)Cao, Zeng, Wang, Sharma, Ma, Liu, and
  Zhou}]{cao2021bert}
Cao, D.; Zeng, K.; Wang, J.; Sharma, P.~K.; Ma, X.; Liu, Y.; and Zhou, S. 2021.
\newblock BERT-Based Deep Spatial-Temporal Network for Taxi Demand Prediction.
\newblock \emph{IEEE Transactions on Intelligent Transportation Systems}.

\bibitem[{Deng et~al.(2021)Deng, Chen, Fan, Jiang, Song, and
  Tsang}]{deng2021pulse}
Deng, J.; Chen, X.; Fan, Z.; Jiang, R.; Song, X.; and Tsang, I.~W. 2021.
\newblock The Pulse of Urban Transport: Exploring the Co-evolving Pattern for
  Spatio-temporal Forecasting.
\newblock \emph{ACM Transactions on Knowledge Discovery from Data (TKDD)},
  15(6): 1--25.

\bibitem[{Deng and Zhang(2021)}]{deng2021comprehensive}
Deng, X.; and Zhang, Z. 2021.
\newblock Comprehensive Knowledge Distillation with Causal Intervention.
\newblock \emph{Advances in Neural Information Processing Systems}, 34.

\bibitem[{Guo et~al.(2021)Guo, Hu, Sun, Qian, Gao, and
  Yin}]{guo2021hierarchical}
Guo, K.; Hu, Y.; Sun, Y.; Qian, S.; Gao, J.; and Yin, B. 2021.
\newblock Hierarchical graph convolution networks for traffic forecasting.
\newblock In \emph{Proceedings of the AAAI Conference on Artificial
  Intelligence}, volume~35, 151--159.

\bibitem[{Han et~al.(2021)Han, Du, Sun, Fu, Lv, and Xiong}]{han2021dynamic}
Han, L.; Du, B.; Sun, L.; Fu, Y.; Lv, Y.; and Xiong, H. 2021.
\newblock Dynamic and Multi-faceted Spatio-temporal Deep Learning for Traffic
  Speed Forecasting.
\newblock In \emph{Proceedings of the 27th ACM SIGKDD Conference on Knowledge
  Discovery \& Data Mining}, 547--555.

\bibitem[{Khemakhem et~al.(2020)Khemakhem, Kingma, Monti, and
  Hyvarinen}]{khemakhem2020variational}
Khemakhem, I.; Kingma, D.; Monti, R.; and Hyvarinen, A. 2020.
\newblock Variational autoencoders and nonlinear ica: A unifying framework.
\newblock In \emph{International Conference on Artificial Intelligence and
  Statistics}, 2207--2217. PMLR.

\bibitem[{Li et~al.(2021{\natexlab{a}})Li, Bai, Liu, Yao, and
  Waller}]{partc2021}
Li, C.; Bai, L.; Liu, W.; Yao, L.; and Waller, S.~T. 2021{\natexlab{a}}.
\newblock A multi-task memory network with knowledge adaptation for multimodal
  demand forecasting.
\newblock \emph{Transportation Research Part C: Emerging Technologies}, 131:
  103352.

\bibitem[{Li et~al.(2021{\natexlab{b}})Li, Feng, Yan, Jin, Jin, and
  Li}]{li2021dynamic}
Li, F.; Feng, J.; Yan, H.; Jin, G.; Jin, D.; and Li, Y. 2021{\natexlab{b}}.
\newblock Dynamic Graph Convolutional Recurrent Network for Traffic Prediction:
  Benchmark and Solution.
\newblock arXiv:2104.14917.

\bibitem[{Li et~al.(2021{\natexlab{c}})Li, Wu, Sun, and Wang}]{li2021causal}
Li, J.; Wu, B.; Sun, X.; and Wang, Y. 2021{\natexlab{c}}.
\newblock Causal Hidden Markov Model for Time Series Disease Forecasting.
\newblock In \emph{Proceedings of the IEEE/CVF Conference on Computer Vision
  and Pattern Recognition}, 12105--12114.

\bibitem[{Liang, Huang, and Zhao(2021)}]{liang2021joint}
Liang, Y.; Huang, G.; and Zhao, Z. 2021.
\newblock Joint Demand Prediction for Multimodal Systems: A Multi-task
  Multi-relational Spatiotemporal Graph Neural Network Approach.
\newblock \emph{arXiv preprint arXiv:2112.08078}.

\bibitem[{Liang et~al.(2021)Liang, Ouyang, Sun, Wang, Zhang, Zheng, Rosenblum,
  and Zimmermann}]{liang2021fine}
Liang, Y.; Ouyang, K.; Sun, J.; Wang, Y.; Zhang, J.; Zheng, Y.; Rosenblum, D.;
  and Zimmermann, R. 2021.
\newblock Fine-Grained Urban Flow Prediction.
\newblock In \emph{Proceedings of the Web Conference 2021}, 1833--1845.

\bibitem[{Liu et~al.(2021{\natexlab{a}})Liu, Sun, Wang, Tang, Li, Qin, Chen,
  and Liu}]{liu2021learning}
Liu, C.; Sun, X.; Wang, J.; Tang, H.; Li, T.; Qin, T.; Chen, W.; and Liu, T.-Y.
  2021{\natexlab{a}}.
\newblock Learning causal semantic representation for out-of-distribution
  prediction.
\newblock \emph{Advances in Neural Information Processing Systems}, 34.

\bibitem[{Liu et~al.(2021{\natexlab{b}})Liu, Wu, Zhuang, Lu, Dou, and
  Xiong}]{liu2021community}
Liu, H.; Wu, Q.; Zhuang, F.; Lu, X.; Dou, D.; and Xiong, H. 2021{\natexlab{b}}.
\newblock Community-Aware Multi-Task Transportation Demand Prediction.
\newblock In \emph{Proceedings of the AAAI Conference on Artificial
  Intelligence}, volume~35, 320--327.

\bibitem[{Locatello et~al.(2019)Locatello, Bauer, Lucic, Raetsch, Gelly,
  Sch{\"o}lkopf, and Bachem}]{locatello2019challenging}
Locatello, F.; Bauer, S.; Lucic, M.; Raetsch, G.; Gelly, S.; Sch{\"o}lkopf, B.;
  and Bachem, O. 2019.
\newblock Challenging common assumptions in the unsupervised learning of
  disentangled representations.
\newblock In \emph{international conference on machine learning}, 4114--4124.
  PMLR.

\bibitem[{Pearl(2009)}]{pearl2009causality}
Pearl, J. 2009.
\newblock \emph{Causality}.
\newblock Cambridge university press.

\bibitem[{Sch{\"o}lkopf(2022)}]{scholkopf2022causality}
Sch{\"o}lkopf, B. 2022.
\newblock Causality for machine learning.
\newblock In \emph{Probabilistic and Causal Inference: The Works of Judea
  Pearl}, 765--804.

\bibitem[{Sch{\"o}lkopf et~al.(2021)Sch{\"o}lkopf, Locatello, Bauer, Ke,
  Kalchbrenner, Goyal, and Bengio}]{comprehensive2021towards}
Sch{\"o}lkopf, B.; Locatello, F.; Bauer, S.; Ke, N.~R.; Kalchbrenner, N.;
  Goyal, A.; and Bengio, Y. 2021.
\newblock Towards causal representation learning.
\newblock \emph{arXiv preprint arXiv:2102.11107}.

\bibitem[{Shen et~al.(2020)Shen, Liu, Dong, Lian, Chen, and
  Zhang}]{shen2020disentangled}
Shen, X.; Liu, F.; Dong, H.; Lian, Q.; Chen, Z.; and Zhang, T. 2020.
\newblock Disentangled generative causal representation learning.
\newblock \emph{arXiv preprint arXiv:2010.02637}.

\bibitem[{Wang et~al.(2021)Wang, Guo, Ouyang, Cheng, Wang, Yu, and
  Liu}]{IOT2021}
Wang, Q.; Guo, B.; Ouyang, Y.; Cheng, L.; Wang, L.; Yu, Z.; and Liu, H. 2021.
\newblock Learning Shared Mobility-aware Knowledge for Multiple Urban Travel
  Demands.
\newblock \emph{IEEE Internet of Things Journal}.

\bibitem[{Wu et~al.(2020)Wu, Pan, Long, Jiang, Chang, and
  Zhang}]{wu2020connecting}
Wu, Z.; Pan, S.; Long, G.; Jiang, J.; Chang, X.; and Zhang, C. 2020.
\newblock Connecting the Dots: Multivariate Time Series Forecasting with Graph
  Neural Networks.
\newblock In \emph{Proceedings of the 26th ACM SIGKDD International Conference
  on Knowledge Discovery \& Data Mining}.

\bibitem[{Wu et~al.(2019)Wu, Pan, Long, Jiang, and Zhang}]{graphwavenet2019}
Wu, Z.; Pan, S.; Long, G.; Jiang, J.; and Zhang, C. 2019.
\newblock Graph WaveNet for Deep Spatial-Temporal Graph Modeling.
\newblock In \emph{Proceedings of the Twenty-Eighth International Joint
  Conference on Artificial Intelligence, {IJCAI} 2019, Macao, China, August
  10-16, 2019}, 1907--1913.

\bibitem[{Yang et~al.(2021)Yang, Liu, Chen, Shen, Hao, and
  Wang}]{yang2021causalvae}
Yang, M.; Liu, F.; Chen, Z.; Shen, X.; Hao, J.; and Wang, J. 2021.
\newblock CausalVAE: Disentangled representation learning via neural structural
  causal models.
\newblock In \emph{Proceedings of the IEEE/CVF Conference on Computer Vision
  and Pattern Recognition}, 9593--9602.

\bibitem[{Ye et~al.(2019)Ye, Sun, Du, Fu, Tong, and Xiong}]{ye2019co}
Ye, J.; Sun, L.; Du, B.; Fu, Y.; Tong, X.; and Xiong, H. 2019.
\newblock Co-prediction of multiple transportation demands based on deep
  spatio-temporal neural network.
\newblock In \emph{Proceedings of the 25th ACM SIGKDD International Conference
  on Knowledge Discovery \& Data Mining}, 305--313.

\bibitem[{Ye et~al.(2021)Ye, Sun, Du, Fu, and Xiong}]{ye2021coupled}
Ye, J.; Sun, L.; Du, B.; Fu, Y.; and Xiong, H. 2021.
\newblock Coupled Layer-wise Graph Convolution for Transportation Demand
  Prediction.
\newblock In \emph{Proceedings of the AAAI Conference on Artificial
  Intelligence}, volume~35, 4617--4625.

\bibitem[{Yu et~al.(2019)Yu, Chen, Gao, and Yu}]{yu2019dag}
Yu, Y.; Chen, J.; Gao, T.; and Yu, M. 2019.
\newblock DAG-GNN: DAG structure learning with graph neural networks.
\newblock In \emph{International Conference on Machine Learning}, 7154--7163.
  PMLR.

\bibitem[{Zhou et~al.(2021)Zhou, Gu, Lu, Zhuang, Zhao, Wang, and
  Zhang}]{zhou2021modeling}
Zhou, Q.; Gu, J.; Lu, X.; Zhuang, F.; Zhao, Y.; Wang, Q.; and Zhang, X. 2021.
\newblock Modeling Heterogeneous Relations across Multiple Modes for Potential
  Crowd Flow Prediction.
\newblock In \emph{Proceedings of the AAAI Conference on Artificial
  Intelligence}, volume~35, 4723--4731.

\end{thebibliography}

\end{document}